\input{Def}
\documentclass[10pt,twocolumn,letterpaper]{article}

\usepackage{cvpr}
\usepackage{times}
\usepackage{epsfig}
\usepackage{graphicx}
\usepackage{amsmath}
\usepackage{amssymb}

\usepackage{rotating}
\usepackage{multirow}
\usepackage{setspace}
\usepackage{algorithm2e}
\usepackage{fancyhdr} 
\usepackage{caption}
 \usepackage{color}
\usepackage{array}
\usepackage{subfigure}
\usepackage{soul}
\usepackage{verbatim} 
\usepackage{paralist}
\usepackage{authblk}

\usepackage[pagebackref=true,breaklinks=true,letterpaper=true,colorlinks,bookmarks=false]{hyperref}

 \cvprfinalcopy 

\def\cvprPaperID{2380} 

\ifcvprfinal\fi
\begin{document}

\title{Dense Relational Captioning: \\ Triple-Stream Networks for Relationship-Based Captioning\vspace{-8mm}}
\renewcommand\Authsep{\quad}
\renewcommand\Authand{\quad}
\renewcommand\Authands{\quad}
\makeatletter
\renewcommand\AB@affilsepx{ \protect\Affilfont}


\author[1]{Dong-Jin Kim}
\author[1]{Jinsoo Choi} 
\author[2]{Tae-Hyun Oh
} 
\author[1]{In So Kweon\vspace{-3mm}}
\affil[1]{KAIST, South Korea.}
\affil[2]{
MIT CSAIL, Cambridge, MA.
}
\affil[1]{\texttt{\{djnjusa,jinsc37,iskweon77\}@kaist.ac.kr }} 
\affil[2]{\texttt{taehyun@csail.mit.edu}}

\maketitle
\thispagestyle{empty}

\begin{abstract}
Our goal in this work is to train an image captioning model that generates more dense and informative captions.
We introduce ``relational captioning,'' a novel {image captioning} task which aims to generate multiple captions with respect to relational information between objects in an image.
Relational captioning is a framework that is advantageous in both diversity and amount of information, leading to image understanding based on relationships.
{Part-of-speech} (POS, \ie subject-object-predicate categories) tags can be assigned to every English word.
We leverage the POS as a prior to guide the correct sequence of words in a caption.
To this end, we propose a multi-task triple-stream network (MTTSNet) which consists of three recurrent units for the respective POS and jointly performs POS prediction and captioning. 
We demonstrate more diverse and richer representations generated by the proposed model against several baselines and competing methods.
The code is available at \href{https://github.com/Dong-JinKim/DenseRelationalCaptioning}{https://github.com/Dong-JinKim/DenseRelationalCaptioning}.
\end{abstract}

\section{Introduction}
Human visual system has the capability to effectively and instantly collect a holistic understanding of contextual associations among objects in a scene~\cite{land2002organization,oliva2007role} by densely and adaptively skimming the visual scene through the eyes, \ie the saccadic movements.
Such instantly extracted rich and dense information allows humans to have the superior capability of object-centric visual understanding.
Motivated by this, in this work, we present a new concept of scene understanding, called \emph{dense relational captioning}, that provides dense but selective, expressive, and relational representation in a human interpretable way, i.e., via captions.



Richer representation of an image often leads to numerous potential applications or performance improvements of subsequent computer vision algorithms~\cite{mottaghi2014role,oliva2007role}. 
In order to achieve richer object-centric understanding, Johnson \etal~\cite{johnson2016densecap} proposed a framework called DenseCap that generates captions for each of the densely sampled local image regions.
{These regional descriptions facilitate both rich and dense semantic understanding of a scene in a form of interpretable language.}
However, the information in the image that we want to acquire includes not only the information of the object itself but also the \emph{interaction} with other objects or the environment.

\begin{figure}[t]
	\centering
		\includegraphics[width=1.0\linewidth,keepaspectratio]{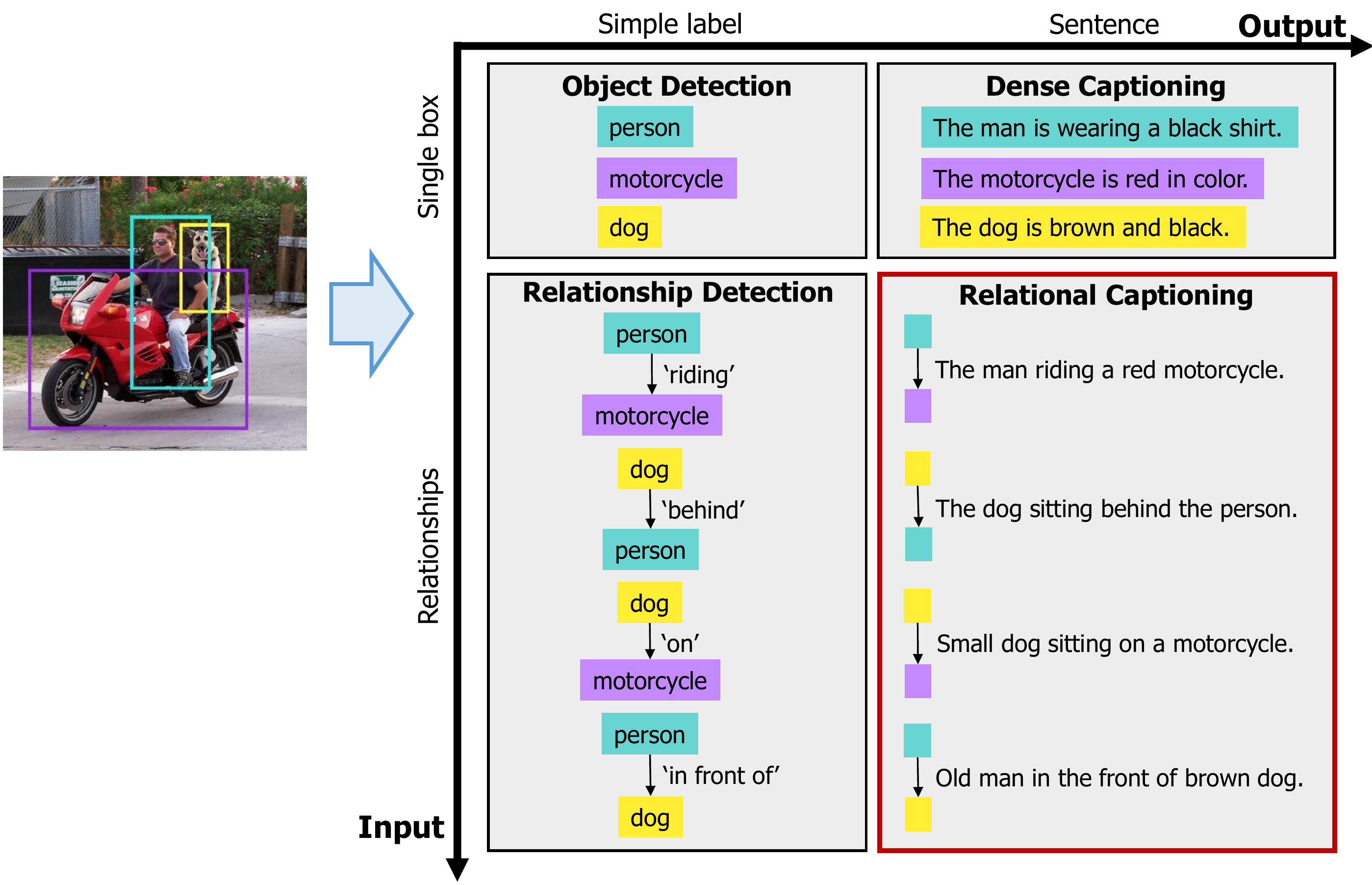}
	\vspace{-7mm}
	\caption{Overall description of the proposed relational captioning framework. Compared to traditional frameworks, our framework is advantageous in both interaction understanding and high-level interpretation.\vspace{-3mm}}
	\label{fig:teaser}
\end{figure}

As {an alternative way} of representing an image, we focus on dense \emph{relationships} between objects.
In the context of human cognition, there has been a general consensus that objects and particular environments near the target object affect search and recognition efficiency.
Understanding the relationships between objects clearly reveal object interactions and object-attribute combinations~\cite{Johnson_2017_ICCV,kim2018disjoint,lu2016visual}.

Interestingly, we observe that the annotations done by humans on computer vision datasets predominantly contain relational forms;
in Visual Genome~\cite{krishna2017visual} and MS COCO~\cite{lin2014microsoft} caption datasets, 
most of the labels take the format of subject-predicate-object more so than subject-predicate.
Moreover, UCF101~\cite{soomro2012ucf101} action recognition dataset contains 85 actions out of 101 (84.2\%) that are described in terms of human interactions with other objects or surroundings. 
These aspects tell us that understanding interaction and relationships between objects facilitate a major component in visual understanding of object-centric events.

In this regard, we introduce a novel captioning framework \emph{relational captioning} that can provide diverse and dense representations from an image.
In this task, we first exploit the relational context between two objects as a representation unit.
This allows generating a combinatorial number of localized regional information.
Secondly, we make use of captioning and its ability to express significantly richer concepts beyond the limited label space of object classes used in object detection tasks.
Due to these aspects, our relational captioning expands the regime further along the label space both in terms of density and complexity, and provides richer representation for an image.


Our main contributions are summarized as follows.
(1) We introduce \emph{relational captioning}, a new {captioning} task that 
generates captions with respect to relational information between objects in an image.
(2) In order to efficiently train the relational caption information, we propose the \emph{multi-task triple-stream network} (MTTSNet) that consists of three recurrent units trained via multi-task learning.
(3) 
{We show that our proposed method is able to generate denser and more diverse captions by evaluating on our relational captioning dataset augmented from Visual Genome (VG)~\cite{krishna2017visual} dataset.}
(4) We introduce {several applications of our framework, including} ``caption graph'' generation which contains richer and more diverse information than conventional scene graphs.

\section{Related Work}
Our work relates to two topics: image captioning and relationship detection.
In this section, we categorize and review related work on these topics.

\noindent{\textbf{Image captioning.}}
By virtue of deep learning and the use of recurrent neural network (\eg LSTM~\cite{hochreiter1997long}) based decoders, image captioning~\cite{ordonez2011im2text} techniques have been extensively explored~\cite{anderson2018bottom,donahue2015long,jiang2018recurrent,karpathy2015deep,lu2016knowing,rennie2017self,vinyals2015show,xu2015show,yao2018exploring,you2016image}.
One of the research issues in captioning is the generation of diverse and informative captions. Thus, learning to generated diverse captions has been extensively studied recently~\cite{chen2018groupcap,dai2017towards,dai2017contrastive,shetty2017speaking,venugopalan2017captioning,wang2017diverse}.
As one of the solutions, the dense captioning (DenseCap) task~\cite{johnson2016densecap} was proposed which uses diverse region proposals to generate localized descriptions, extending the conventional holistic image captioning to diverse captioning that can describe local contexts. 
{Moreover, our \emph{relational captioning} is able to generate even more diverse caption proposals than dense captioning by considering \emph{relations} between objects.}

Yang~\etal\cite{Yang_2017_CVPR} improves the DenseCap model by incorporating a global image feature as context cue as well as a region feature of the desired objects with a late fusion.
Motivated by this, 
in order to implicitly learn dependencies of subject, object and union representations, we incorporate a triple-stream LSTM for our captioning module.

\begin{figure*}[t]
\vspace{-2mm}\begin{center}		\includegraphics[width=1.0\linewidth,keepaspectratio]{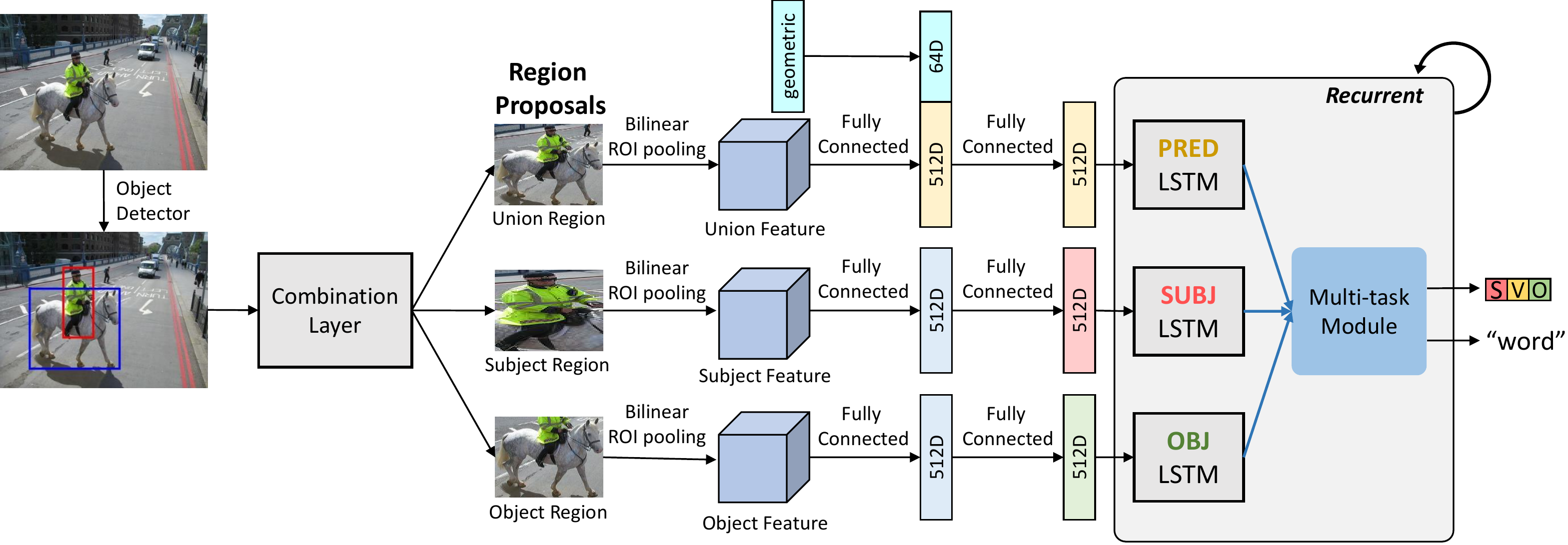}
	\end{center}
	\vspace{-6mm}
	\caption{Overall architecture of the proposed multi-task triple-stream networks. 
    Three region features (Union, Subject, Object) come from the same shared branch (Region Proposal Network), and for \emph{subject} and \emph{object} features, the first intermediate FC layer share weights (depicted in the same color).
    \vspace{-3mm}
    }
	\label{fig:architecture}
\end{figure*}

\noindent{\textbf{Visual relationship detection (VRD).}}
Understanding visual relationships between objects have been an important concept in various tasks.
Conventional VRD usually deals with predicting the subject-predicate-object (in short, \texttt{subj-pred-obj}).
A pioneering work by Lu~\etal\cite{lu2016visual} formalizes the VRD task and provides a dataset, while addressing the subject (or object) and predicate classification models separately.
{On the other hand, similar to VRD task, scene graph generation (a task to generate a structured graph that contains the context of a scene) has also started to be explored~\cite{li2017scene,woo2018linknet,xu2017scene,zellers2018neural}. }

Although the VRD dataset is larger ($100$ object classes and $70$ predicates) than Visual Phrases, it is still inadequate to handle the real world scale.
The Visual Genome (VG) dataset~\cite{krishna2017visual} for relationship detection consists of $31$k predicate types and $64$k object types giving the number of possible combinations of relationship triplets too diverse for the state-of-the-art VRD based models.
{This is because the labels consist of the various combinations of words (\eg `little boy,' `small boy,' \etc)}
As a result, only the simplified version of VG relationship dataset has been studied. 
On the contrary, our method is able to generate relational captions by tokenizing the whole relational expressions into words, and learning from them.

While the recent state-of-the-art VRD~\cite{li2017vip,lu2016visual,plummer2016phrase,yu2017visual,yin2018zoom} {or scene graph generation works~\cite{li2017scene,woo2018linknet,xu2017scene,zellers2018neural}} mostly use language priors to detect relationships, we directly learn the relationship as a descriptive language model.
{In addition, the expressions of traditional scene graph generation or VRD task are restricted to \texttt{subj-pred-obj} triplets, whereas the relational captioning is able to provide additional information such as attributes or noun modifiers by adopting free-form natural language expressions. }

In summary, dense captioning facilitates a natural language interpretation of regions in an image, while VRD can obtain relational information between objects. 
Our work combines both axes, resulting in much denser and diverse captions than DenseCap.
That is, given $B$ region proposals in an image, we can obtain $B(B{-}1)$ relational captions, whereas DenseCap returns only $B$ captions. 

\section{Multi-task Triple-Stream Networks}

Our relational captioning is defined as follows. 
Given an input image, a bounding box detector generates various object proposals and a captioning module predicts combinatorial captions with POS labels describing each pair of objects.
Figure~\ref{fig:architecture} shows the overall framework of the proposed relational captioning model, which is mainly composed of a localization module based on the region proposal network (RPN)~\cite{ren2015faster}, and a triple-stream RNN (LSTM~\cite{hochreiter1997long}) module for captioning.
Our network supports end-to-end training with a single optimization step that allows joint localization, combination, and description with natural language.

Given an image, RPN generates object proposals.
Then, the combination layer takes a pair consisting of a \emph{subject} and an \emph{object} at a time.
To take the surrounding context information into account, we utilize the \emph{union} region of the \emph{subject} and \emph{object} regions, in a way similar to using the global image region as side information by Yang~\etal\cite{Yang_2017_CVPR}.
This feature of triplets (\emph{subject}, \emph{object}, \emph{union}) are fed to the triple-stream LSTMs, where each stream takes its own purpose, \ie\emph{subject}, \emph{object}, and \emph{union}.
Given this triplet feature, the triple-stream LSTMs collaboratively generate a caption and POS classes of each word.
We describe these processes as follows.

\subsection{Region Proposal Networks}
Our network uses fully convolutional layers of VGG-16~\cite{simonyan2014very} up to the final pooling layer (\ie\texttt{conv5\_3}) for extracting the spatial features via the bilinear ROI pooling~\cite{johnson2016densecap}.
The object proposals are generated by localization layers.
It takes the feature tensor, and proposes $B$ regions (user parameter) of interest.
Each proposed region has its confidence score, region feature of shape $512{\times}7{\times}7$, and coordinates $b {=} (x,y,w,h)$ of the bounding box with center $(x,y)$, width $w$ and height $h$.
We process it into vectorized features (of shape $D{=}512$) using two fully-connected (FC) layers. 
This encodes the appearance of each region into a feature, called \emph{region code}.
Once the region codes are extracted, they are reused for the following processes.

To generate relational proposals, we build pairwise combinations of $B$ region proposals, where in turn we get $B(B{-}1)$ possible region pairs.
We call this layer the \emph{combination layer}.
A distinctive point of our model with the previous dense captioning works~\cite{johnson2016densecap,Yang_2017_CVPR}, is that while the works regard each region proposal as an independent target to describe and produce $B$ number of captions, we consider their pairwise combinations $B(B{-}1)$, which are much denser and explicitly expressible in term of relationships. 
Also, we can asymmetrically use each entry of a pair by assigning the roles of the regions, \ie (\emph{subject}, \emph{object}).

\begin{figure*}[t]
	\centering
	\includegraphics[width=1.0\linewidth,keepaspectratio]{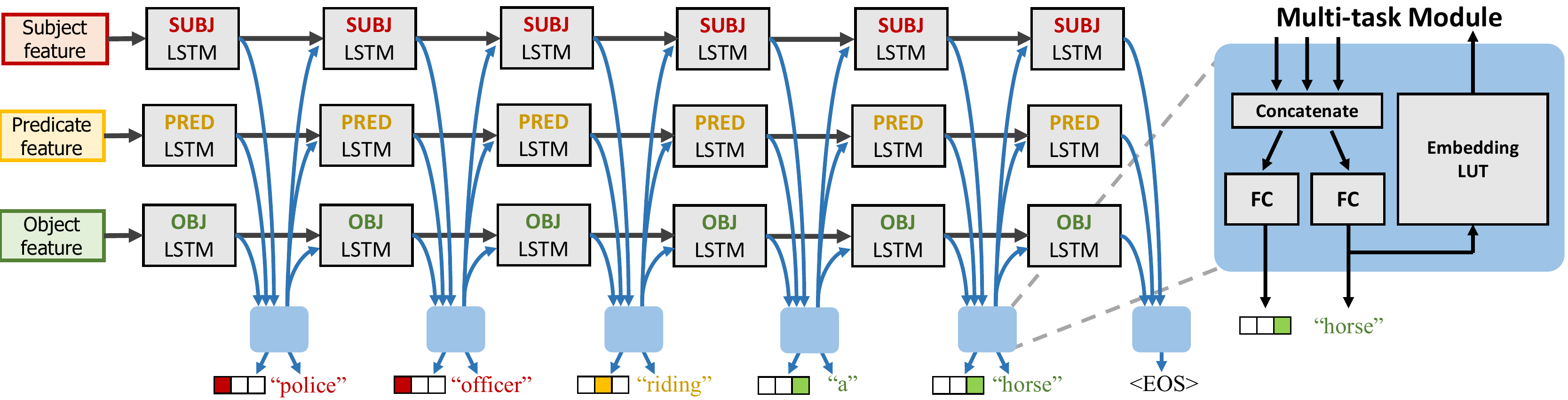}
	\caption{An illustration of the unrolled triple-stream LSTM. 
		Our model consists of two major parts: triple-stream LSTM and a multi-task module.
		The multi-task module jointly predicts a caption word and its POS class (\texttt{subj-pred-obj}, illustrated as three cells colored according to the POS class), as well as the input vector for the next time step.
        \vspace{-4mm}}
	\label{fig:triple-stream}
\end{figure*}

Furthermore, motivated by Yang~\etal, where the global context of an image improves the captioning performance, we leverage an additional region, the \emph{union} region $b_u$ of (\emph{subject}, \emph{object}).
In addition, to provide relative spatial information, we append a geometric feature for the \emph{subject} and \emph{object} box pair, \ie$(b_s, b_o)$ to the \emph{union} feature before the FC layers.
Given two bounding boxes $b_s$ and $b_o$, the geometric feature $\mathbf{r}$ is defined similarly to \cite{peyre2017weakly} as\vspace{-1mm}
\begin{equation}
        \mathbf{r} = \left[ \tfrac{x_o - x_s}{\sqrt{w_s h_s}} ,  \tfrac{y_o - y_s}{\sqrt{w_s h_s}}, \sqrt{\tfrac{w_o h_o}{w_s h_s}} , 
		\tfrac{w_s}{h_s}, \tfrac{w_o}{h_o} , 
		\tfrac{b_s \bigcap b_o}{b_s \bigcup b_o} \right] \in \R^6.\vspace{-1mm}
		\label{eq:geometric}
\end{equation}
By concatenating the \emph{union} feature with $\mathbf{r}$ which is passed through an additional FC layer, the shape of this feature is $D{+}64$.
Then, the dimension of the \emph{union} region code is reduced by the following FC layers.
This stream of operations is illustrated in \Fref{fig:architecture}.
The three features extracted from the \emph{subject}, \emph{object}, and \emph{union} regions are fed to each LSTM described in the following sections.

\subsection{Relational Captioning Networks }
Relational caption generation takes the {relational information of the object pairs} into account.
{However, expressing the relationship in a sentence has been barely studied.}
Therefore, we design a new network that deals with relational captions, called the \emph{multi-task triple-stream network}.

From the region proposal network, a triplet of region codes are fed as input to LSTM cells, so that a sequence of words (caption) is generated.
In the proposed relational region proposal, a distinctive facet is to provide a triplet of region codes consisting of \emph{subject}, \emph{object}, and \emph{union} regions, which virtually corresponds to the POS of a sentence  (\texttt{subj}-\texttt{pred}-\texttt{obj}).
This correspondence between regions in a triplet and POS information leads to the following advantages: 1) input features can be {adaptively merged} depending on its POS and fed to the caption generation module, and 2) the POS prior on predicting a word can be effectively applied to caption generation.
However, leveraging and processing these input cues are non-trivial.

For the first advantage, in order to derive POS aware inference, we propose \emph{triple-stream networks}, which are three separate LSTMs respectively corresponding to \texttt{subj}-\texttt{pred}-\texttt{obj}. 
The outcomes of LSTMs are combined via concatenation.
For the second advantage, during a word prediction, we jointly infer its POS class via multi-task inference.
This POS class prediction acts as a prior for the word prediction of a caption during the learning phase.

\vspace{1mm}\noindent\textbf{Triple-Stream LSTMs.}\quad
Intuitively, the region codes of \emph{subject} and \emph{object} would be closely related to the subject and object related words in a caption, while the \emph{union} and geometric features may contribute to the predicate.
In our relational captioning framework, the LSTM modules must adaptively take input features into account according to which POS decoding stage it is for a caption.

As shown in Fig.~\ref{fig:architecture}, the proposed triple-stream LSTM module consists of three separate LSTMs, each of which is in charge of the \emph{subject}, \emph{object} and \emph{union} region codes respectively.
At each step, the triple-stream LSTMs generate three embedded representations separately, and a single word is predicted by consolidating the three processed representations.
The embedding of the predicted word is distributed into all three LSTMs as inputs and is used to run the next step in a recursive manner.
Thus in each step, each entry of the triplet input is used differently, which allows more flexibility than a single LSTM as used in traditional captioning models~\cite{johnson2016densecap,vinyals2015show}.
In other words, the weights of the input cue features change at every recursive step according to which POS the word being generated belongs to.

\vspace{1mm}\noindent\textbf{Multi-task with POS Classification.}\quad
On top of this concatenation, we utilize the POS information to more effectively train the relational captioning model. 
Relational captioning generates a sequence of words in \texttt{subj}-\texttt{pred}-\texttt{obj} order, \ie the order of POS.
For each word prediction, in a \emph{multi-task} module in Fig.~\ref{fig:triple-stream}, we also classify the POS class of the predicted word, so that it encourages the caption generation to follow the word order in the POS order.

When three representations {for each POS} are to be consolidated, one option can be to consolidate them in an early step, called \emph{early fusion}.
This results in a single LSTM with the fusion of the three region codes (\eg concatenation of three codes).
However, as reported by Yang~\etal\cite{Yang_2017_CVPR}, this early fusion approach also shows lower performance than that of late fusion methods.
In this regard, we adopt a \emph{late fusion} for a multi-task module.
The layer basically concatenates the representation outputs from the triple-stream LSTMs, but due to the recurrent multi-task modules, it is able to generate sophisticated representations.

We empirically observe that this multi-task learning with POS helps not only the shared representation to become richer but also guides the  word predictions, and thus helps to improve the captioning performance overall. 
We hypothesize that the POS task provides distinctive information that may help learn proper representations from the triple-stream LSTMs.
Since each POS class prediction tightly relies on respective representations from each LSTM, \eg \emph{pred}-LSTM closely related to \texttt{pred} of POS, the gradients generated from the POS classification would be {back-propagated} through the indices of the concatenated representation according to the class.
By virtue of this, the multi-task triple-stream LSTMs are able to learn the representation in such a way that it can predict plausible words for each time step.
{Therefore,} our model can generate appropriate words according to the POS at a given time step.

\noindent\textbf{Loss functions.}\quad
Training our relational captioning model can be mainly divided into captioning loss and detection loss. 
Specifically, the proposed model is trained to minimize the following loss function:
\begin{equation}
		\mathcal{L} = \mathcal{L}_{cap} + \alpha \mathcal{L}_{POS} + \beta \mathcal{L}_{det} + \gamma \mathcal{L}_{box},
		\label{eq:loss}
\end{equation}
where $\mathcal{L}_{cap}$, $\mathcal{L}_{POS}$, $\mathcal{L}_{det}$, and $\mathcal{L}_{box}$ denote captioning loss, POS classification loss, detection loss, and bounding box regression loss respectively.
$\alpha$, $\beta$, and $\gamma$ are the balance parameters (we set them to $0.1$ for all experiments). 

The first two terms are for captioning and the next two terms are for the region proposal.
$\mathcal{L}_{cap}$ and $\mathcal{L}_{POS}$ are cross-entropy losses at every time step for each word and POS classification respectively. 
For each time step, $\mathcal{L}_{POS}$ measures a 3-class cross entropy loss.
$\mathcal{L}_{det}$ is a binary logistic loss for foreground/background regions, while $\mathcal{L}_{box}$ is a smoothed L1 loss ~\cite{ren2015faster}. 

\section{Experiments}

In this section, we provide the experimental setups, competing methods and performance evaluation of relational captioning with both quantitative and qualitative results.
\subsection{Relational Captioning Dataset}

{Since there is no existing dataset for the relational captioning task, we construct a dataset by utilizing VG relationship dataset version 1.2~\cite{krishna2017visual} which consists of 85200 images with 75456/4871/4873 splits for train/validation/test sets respectively.
We tokenize the relational expressions to form natural language expressions, and for each word, we assign the POS class from the triplet association.
}

{
However, VG relationship datasets show limited diversity in the words used.
Therefore, by only using relational expressions to construct data, the captions generated from a model tends to be simple (\emph{e.g.} ``building-has-window'').
{Even though our model may enable richer concepts and expressions,} if the training data does not contain such concepts and expressions, there is no way to actually \emph{see} this.
In order to validate the diversity of our relational captioner, we need to make our relational captioning dataset to have more natural sentences with rich expressions.}

Through observation, {we noticed that the relationship dataset labels lack \emph{attributes} describing the subject and object, which are perhaps what enriches the sentences the most}.
Therefore, we utilize the \emph{attribute labels} of VG data to augment existing relationship expressions.
More specifically, we simply find the attribute that matches the subject/object of the relationship label and attach it to the subj/obj caption label.
In particular, if an attribute label describes the same subject/object for a relationship label while associated bounding box overlaps enough, the label is considered to be \emph{matched} to the subject/object in the relationship label.
{After this process, we obtain 15595 vocabularies for our relational captioning dataset (11447 vocabularies before this process).}
{We train our caption model with this data, and report its result in this section.}
In addition, we provide a holistic image captioning performance and various analysis such as comparison with scene graph generation.

\begin{table}[t]
\vspace{0mm}
    \centering
    \resizebox{1.0\linewidth}{!}{%
         \begin{tabular}{l|| c| c|c}\hline
										&	mAP (\%) 	&Img-Lv. Recall	&METEOR\\\hline\hline
            \texttt{Direct Union}		&		--	&	17.32	&11.02\\\hline
            \texttt{Union}				&	0.57	&	25.61	&12.28\\
            \texttt{Union+Coord.}		&	0.56 	&	27.14 	&13.71\\
            \texttt{Subj+Obj}			&	0.51 	& 	28.53	&13.32\\
           \texttt{Subj+Obj+Coord.}		&	0.57 	&	30.53 	&14.85\\
           \texttt{Subj+Obj+Union}		& 0.59		& 	30.48	& 15.21\\
           \texttt{\textbf{TSNet (Ours)} }	&	\textbf{0.61}	& 	\textbf{32.36}		&\textbf{16.09}\\
           \hline
           \texttt{Union (w/MTL)}		&	0.61	&	26.97 	&12.75\\
           \texttt{Subj+Obj+Coord (w/MTL)}		&	0.63 	&	31.15 	&15.31\\
           \texttt{Subj+Obj+Union (w/MTL)}		& 0.64	& 	31.63		& 16.63\\
            \texttt{\textbf{MTTSNet (Ours)} } &\textbf{0.88}&\textbf{34.27}	&\textbf{18.73}\\
            \hline
            \hline
            \texttt{Neural Motifs~\cite{zellers2018neural} } &0.25 & 29.90&15.34\\
            \hline
            
		\end{tabular}
        
        }
        \vspace{-2mm}
	\caption{Ablation study for relational dense captioning task on relational captioning dataset. 
                \vspace{-4mm} }
	\label{table:captioning}	
\end{table}

 \begin{figure*}[t]
\centering
\resizebox{1\linewidth}{!}{%
{\includegraphics[height=0.5\linewidth,keepaspectratio]{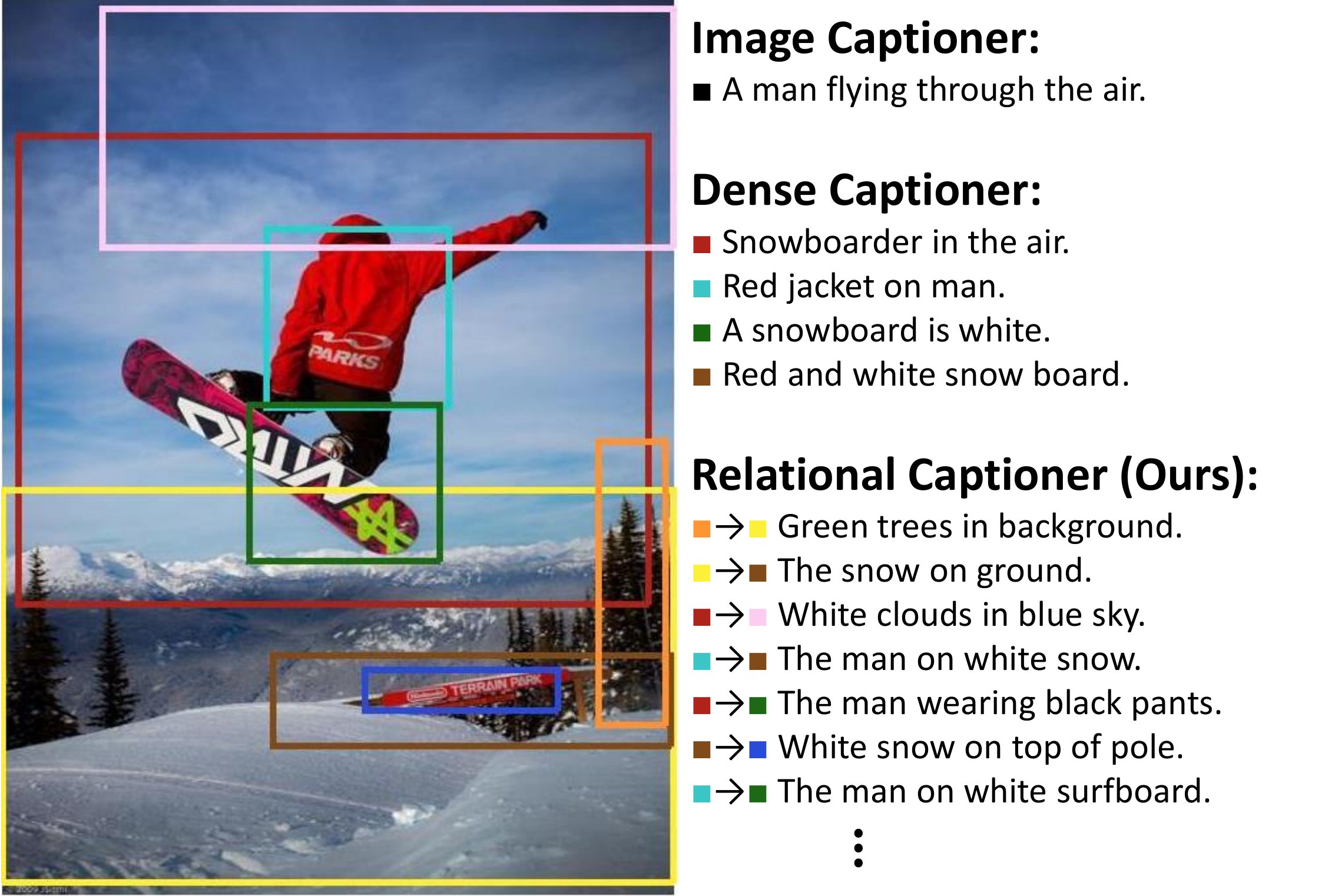}}
{\includegraphics[height=0.5\linewidth,keepaspectratio]{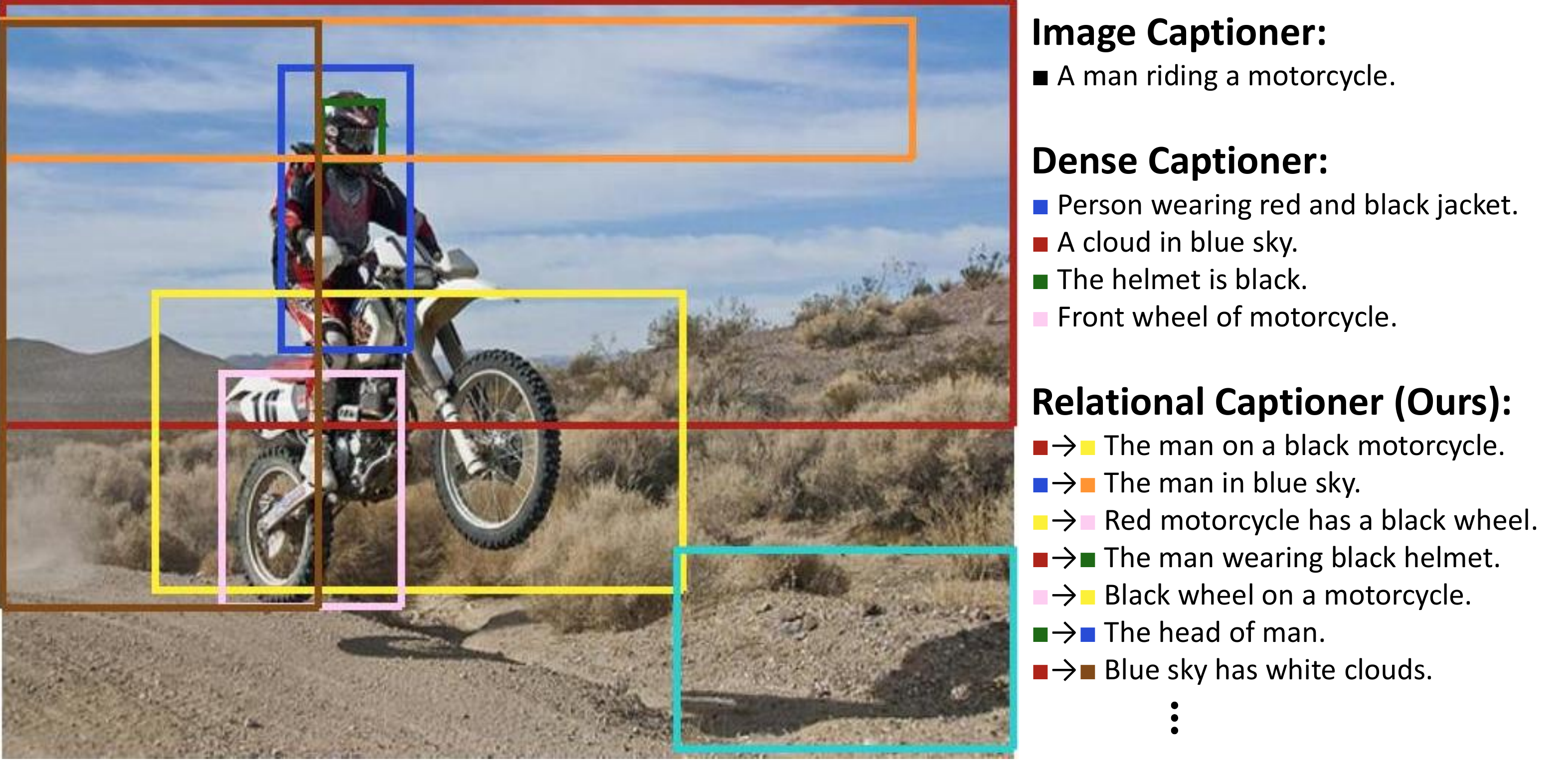}}}\\
\vspace{-2mm}
\caption{Example captions and region generated by the proposed model. 
   We compare our result with the image captioner~\cite{vinyals2015show} and the dense captioner~\cite{johnson2016densecap} in order to contrast the amount of information and diversity. 
   \vspace{-2mm}}
\label{fig:qualitative}
\end{figure*}

\begin{table*}[t]
\centering
    \resizebox{0.6\linewidth}{!}{%
		\begin{tabular}{l|| c|c||c|c}\hline
													&		Recall	&	METEOR		&\#Caption	&Caption/Box	\\\hline\hline
			Image Captioner (Show\&Tell)~\cite{vinyals2015show} 	&		23.55	&	8.66		&	1		&	1		\\
            Image Captioner (SCST)~\cite{rennie2017self} 	&		24.04	&	14.00		&	1		&	1		\\
			Dense Captioner (DenseCap)~\cite{johnson2016densecap}  &		42.63	&	19.57	&		9.16	&	1		\\\hline
			Relational Captioner (\texttt{Union})	&		38.88	&	18.22	&	85.84	&	9.18	\\
			Relational Captioner (\textbf{\texttt{MTTSNet}})		&\textbf{46.78} &\textbf{21.87}	&\textbf{89.32}&\textbf{9.36}\\\hline
		\end{tabular}
    }
    \vspace{-2mm}
	\caption{Comparisons of the holistic level image captioning. We compare the results of the relational captioners with that of two image captioners~\cite{rennie2017self,vinyals2015show} and a dense captioner~\cite{johnson2016densecap}. \vspace{-4mm}}
	\label{table:recall}	
\end{table*}


\begin{figure*}[t]
\centering
\resizebox{1\linewidth}{!}{%
    \begin{tabular}[c]{c}
    \subfigure[]{\includegraphics[width=0.35\linewidth,keepaspectratio]{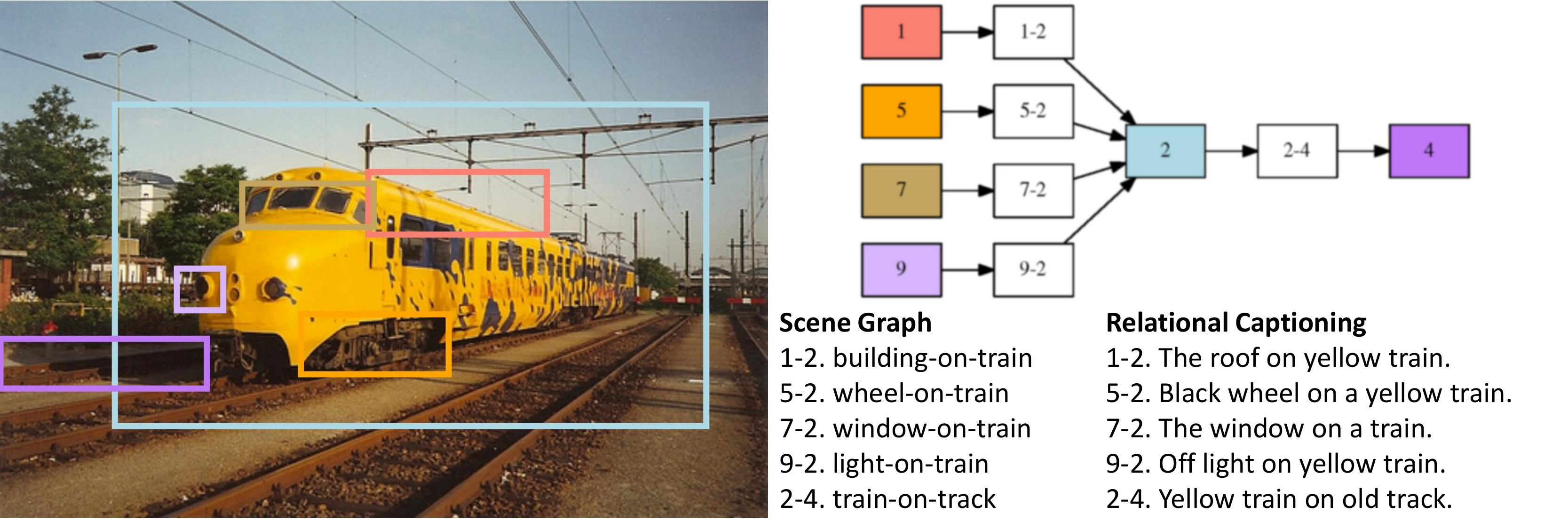}}				
    \subfigure[]{\includegraphics[width=0.35\linewidth,keepaspectratio]{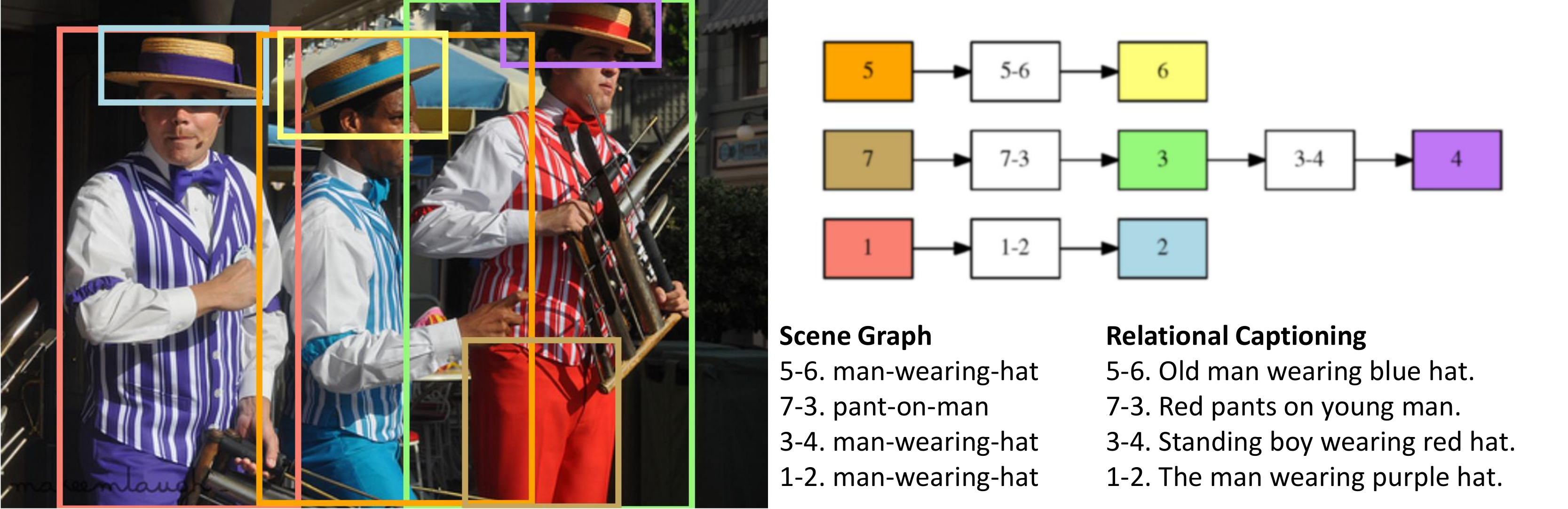}}\\
    \subfigure[]{\includegraphics[width=0.35\linewidth,keepaspectratio]{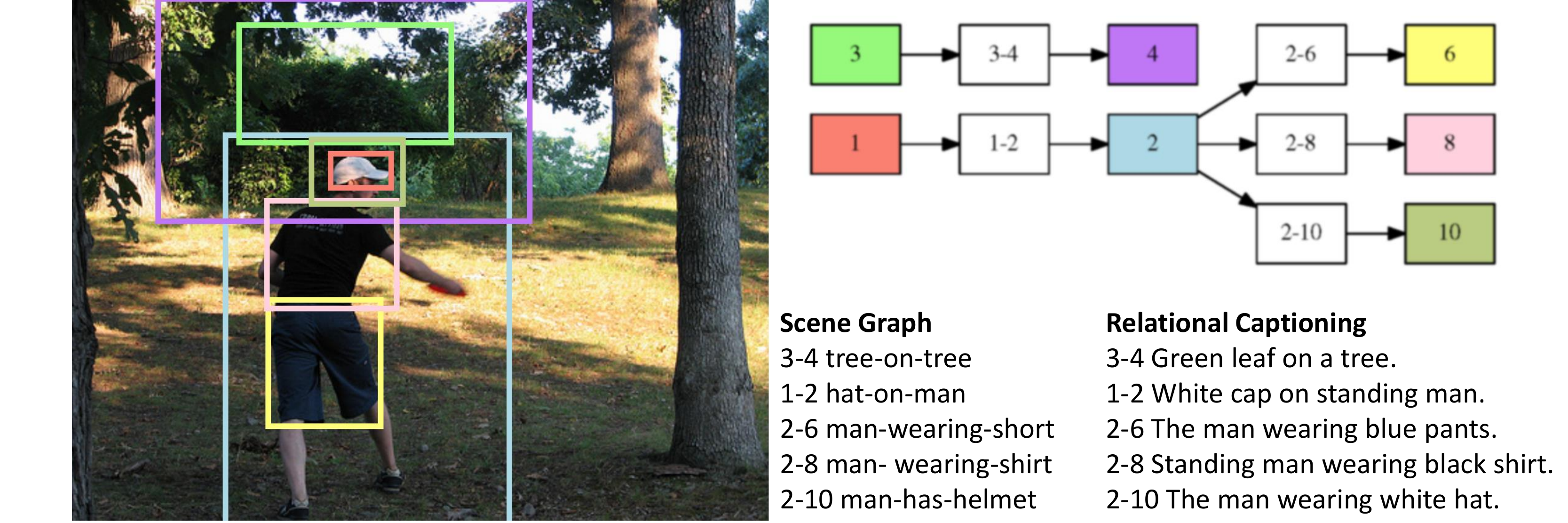}}
    \subfigure[]{\includegraphics[width=0.35\linewidth,keepaspectratio]{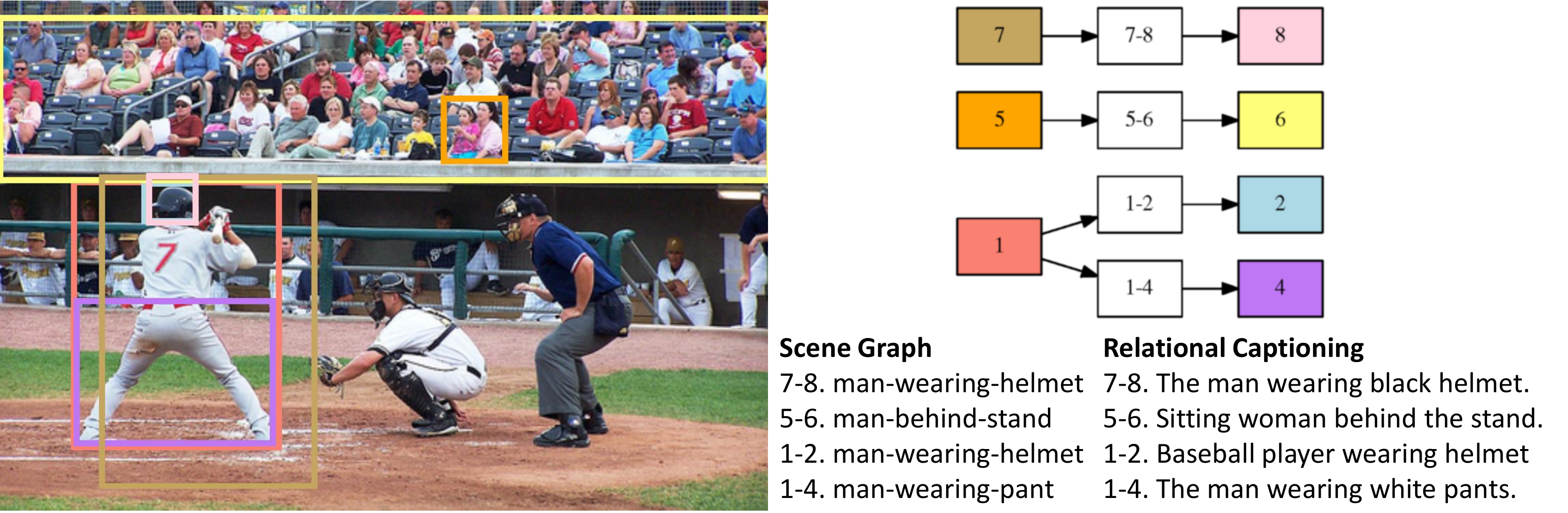}}\\
    \end{tabular}
}
	\vspace{-4mm}
	\caption{ Results of generating ``caption graph'' from our relational captioniner. In order to compare the diversity of the outputs, we also show the result of the scene graph generator, Neural Motifs~\cite{zellers2018neural}.\vspace{-4mm}} 
	\label{fig:scene-graph}
\end{figure*}

\subsection{Relational Dense Captioning: Ablation Study}
\label{sec:exp_relation}

\noindent\textbf{Baselines.} Since no direct work for relational captioning exists, we implement several baselines by modifying the most relevant methods, which facilitate our ablation study.

{\setdefaultleftmargin{0mm}{}{}{}{}{}
\begin{itemize}
\item \texttt{Direct Union} has the same architecture with \emph{DenseCap}~\cite{johnson2016densecap}, but of which RPN is trained to directly predict union regions.
The union region is used to generate captions by one LSTM.\vspace{0mm}

\item \texttt{Union} also resembles \emph{DenseCap}~\cite{johnson2016densecap} and \texttt{Direct union}, but its RPN predicts individual object regions. 
The object regions are paired as (subject, object), and then a union region from each pair is fed to a single LSTM for captioning.
Also, we implement two additional variants: \texttt{Union (w/MTL)} additionally predicts the POS classification task, and 
\texttt{Union+Coord.} appends the geometric feature to the region code of the union.\vspace{0mm}

\item \texttt{Subj+Obj} and \texttt{Subj+Obj+Union} models use the concatenated region features of (subject, object) and (subject, object, union) respectively and pass them through a single LSTM (early fusion approach).
Also, \texttt{Subj+Obj+Coord.} uses the geometric feature instead of the region code of the union.
Moreover, we {evaluate the baselines}, \texttt{Subj+Obj+\{Union,Coord\}} with POS classification (MTL loss).
\vspace{0mm}

\item \texttt{TSNet} denotes the proposed triple-stream LSTM based model without a branch for POS classifier.
Each stream takes the region codes of (subject, object, union + coord.) separately. 
\texttt{MTTSNet} denotes our final model, multi-task triple-stream network with POS classifier.

\end{itemize}
}

\noindent\textbf{Evaluation metrics.}
Motivated by the evaluation metric suggested for dense captioning task~\cite{johnson2016densecap},
we suggest a new evaluation metric for relational dense captioning.
We report the mean Average Precision (mAP) which measures both localization and language accuracy. 
As suggested by Johnson~\etal, we use METEOR score~\cite{denkowski2014meteor} with thresholds $\{ 0, 0.05, 0.1 0.15, 0.2, 0.25\}$ for language, and IOU thresholds $\{0.2, 0.3, 0.4, 0.5, 0.6\}$ for localization.
The AP values obtained by all the pairwise combinations of language and localization thresholds are averaged to get the final mAP score.
The major difference of our metric is that, for the localization AP, we measure for both the subject and object bounding boxes with respective ground truths. 
{In particular, we only consider the samples with IOUs of both the subject and object bounding boxes greater than the localization threshold.}
For all cases, we use percentage as the unit of metric.
In addition, we suggest another metric, called ``image-level (Img-Lv.) recall.'' 
This measures the caption quality at the holistic image level by considering the bag of all captions generated from an image as a single prediction.
Given {only} the aforementioned language thresholds for METEOR {\ie without box IOU threshold}, we measure the recall of the predicted captions.
The metric evaluates the diversity of the produced representations by the model for a given image.
Also, we measure the average METEOR score for predicted captions to evaluate the caption quality.

\noindent\textbf{Results.}
Table~\ref{table:captioning} shows the performance of the relational dense captioning task on relational captioning dataset. 
{The second and third row sections (2-7 and 8-11th rows) show the comparison of the baselines with and without POS classification (\texttt{w/MTL})}.
{In the last row, we show the performance of the state-of-the-art scene graph generator, Neural Motifs~\cite{zellers2018neural}.
Due to the different output structure, we compare with Neural Motifs trained with the supervision for relationship detection.}
Similar to the setup in DenseCap~\cite{johnson2016densecap}, we fix the number of region proposals before NMS to 50 for all methods for a fair comparison.

{Among the results in the second row section (2-7th rows) of Table~\ref{table:captioning}}, our TSNet shows {the best result suggesting that the triple-stream component alone is a sufficiently strong baseline over the others.}
On top of TSNet, applying the MTL loss (\ie, MTTSNet) improves overall performance, and especially improves mAP, where the detection accuracy seems to be dominantly improved compared to the improvement of the other metrics.
This shows that \emph{triple-stream LSTM} is the key module that most leverages the MTL loss across other early fusion approaches (see the third {row section} of the table).
As another factor, {we can see from Table~\ref{table:captioning} that the relative spatial information (\texttt{Coord.}) and union feature information (\texttt{Union}) improves the results.} 
This is because 
the union feature itself preserves the spatial information to some extent from the $7\times 7$ grid form of its activation. 
For \texttt{Neural Motifs}, other relational captioner baselines including our \texttt{TSNet} and \texttt{MTTSNet} perform favorably against \texttt{Neural Motifs} in all metrics.
This is worth noting because handling free-form language generation which we aim to achieve is more challenging than the simple triplet prediction of scene graph generation.




\subsection{Holistic Image Captioning Comparison}
We also compare our approach with other image captioning frameworks, \emph{Image Captioner} {(Show\&Tell~\cite{vinyals2015show} and SCST~\cite{rennie2017self})}, and \emph{Dense Captioner}~(DenseCap~\cite{johnson2016densecap}), in a holistic image description perspective.
In order to measure the performance of \emph{holistic image-level} captioning for dense captioning methods, we use Img-Lv. Recall metric {defined in the previous section} (Recall).
We compare them with two relational dense captioning methods, \texttt{Union} and \texttt{MTTSNet}, denoted as \emph{Relational Captioner}.
For a fair comparison, for \emph{Dense} and \emph{Relational Captioner}, we adjust the number of region proposals after NMS to be similar, which is different from the setting in the previous section which fixed the number of proposals before NMS.

Table~\ref{table:recall} shows the image-level recall, METEOR, and additional quantities for comparison. 
\emph{\#Caption} denotes the average number of captions generated from an input image and \emph{Caption/Box} denotes the average ratio of the number of captions generated and the number of boxes remaining after NMS.
Therefore, \emph{Caption/Box} demonstrates how many captions can be generated given the same number of boxes generated after NMS.
By virtue of multiple captions per image from multiple boxes, the \emph{Dense Captioner} is able to achieve higher performance than {both of the} \emph{Image Captioner}s. 
Compared with the \emph{Dense Captioner}, \texttt{MTTSNet} as a \emph{Relational Captioner} can generate an even larger number of captions given the same number of boxes.
Hence, as a result of learning to generate diverse captions, the \texttt{MTTSNet} achieves higher recall and METEOR.
From the performance of \texttt{Union}, we can see that it is difficult to obtain better captions than \emph{Dense Captioner} by only learning to use the union of subject and object boxes, despite having a larger number of captions.

We show example predictions of our relational captioning model in Fig.~\ref{fig:qualitative}. 
Our model is able to generate rich and diverse captions for an image. 
We also show a comparison with the traditional frameworks, image captioner~\cite{vinyals2015show} and dense captioner~\cite{johnson2016densecap}.
While the dense captioner is able to generate diverse descriptions than an image captioner by virtue of various regions, our model can generate an even greater number of captions from the combination of the bounding boxes.

\subsection{Comparison with Scene Graph}
{
Motivated by scene graph, which is derived from the VRD task,
we extend to a new type of a scene graph, which we call ``caption graph.'' 
Figure~\ref{fig:scene-graph} shows the caption graphs generated from our MTTSNet as well as the scene graphs from Neural Motifs~\cite{zellers2018neural}.
{For caption graph, we follow the same procedure as Neural Motifs but replace the relationship detection network into our MTTSNet.}
In both methods, we use ground truth bounding boxes {to generate scene (and caption) graphs} for fair comparison.

By virtue of being free form, our caption graph can have richer expression and information including attributes, whereas the traditional scene graph is limited to a closed set of the \texttt{subj-pred-obj} triplet.
For example, in Fig.~\ref{fig:scene-graph}-(b,d), given the same object `person,' our model is able to distinguish the fine-grained category (\ie man vs boy and man vs woman).
In addition, our model can provide more status information about the object (\eg standing, black), by virtue of the attribute contained in our relational captioning data.
Most importantly, the scene graph can contain unnatural relationships (\eg tree-on-tree in Fig.~\ref{fig:scene-graph}-(c)), because prior relationship detection methods, \eg \cite{zellers2018neural}, predict object classes individually.
In contrast, by predicting the full sentence for every object pair, relational captioner can assign a more appropriate word for an object by considering the relations, \eg ``Green leaf on a tree.''

Lastly, our model is able to assign different words for the same object by considering the context (the man vs baseball player in Fig.~\ref{fig:scene-graph}-(d)), whereas the scene graph generator can only assign one most likely class (man).
Thus, our relational captioning framework enables more diverse interpretation of the objects compared to the traditional scene graph generation models.
}

\begin{table}[t]
\centering
    	\resizebox{.8\linewidth}{!}{%
        \begin{tabular}{l|| c|c}\hline
													&			words/img		&	words/box	\\\hline\hline
			Image Cap.~\cite{vinyals2015show}	&				4.16		&	-	\\
            Scene Graph~\cite{zellers2018neural}				&		7.66		&	3.29	\\
            Dense Cap.~\cite{johnson2016densecap}		&				18.41		&	4.59	\\
            Relational Cap. (\textbf{\texttt{MTTSNet}})	 			&\textbf{20.45}&\textbf{15.31}\\\hline
		\end{tabular}
        }\vspace{0mm}
	\caption{Diversity comparison between image captioning, scene graph generation, dense captioning, and relational captioning.
    We measure the number of different words per image (words/img) and the number of words per bounding box (words/box).\vspace{-4mm}}
	\label{table:diversity}	
\end{table}

\subsection{Additional Analysis}

\noindent\textbf{Vocabulary Statistics.}
In addition, we measure the vocabulary statistics and compare them among the frameworks.
The types of statistics measured are as follows:
1) an average number of unique words that have been used to describe an image, and
2) an average number of words to describe each box.
More specifically, we count the number of unique words in all the predicted sentences and present the average number per image or box.
Thus, the metric measures the amount of information we can obtain given an image or a fixed number of boxes.
The comparison is depicted in Table~\ref{table:diversity}.
These statistics increase from \emph{Image Cap.} to \emph{Scene Graph} to \emph{Dense Cap.} to \emph{Relational Cap.}
{In conclusion, the proposed relational captioning is advantageous in diversity and amount of information, compared to both of the traditional object-centric scene understanding frameworks, scene graph generation and dense captioning.}

\noindent\textbf{Sentence-based Image and Region-pair Retrieval.}
Since our relational captioning framework produces richer image representations than other frameworks, it may have benefits on the sentence based image or region-pair retrieval, \emph{which {cannot} be performed by scene graph generation or VRD models}.
To evaluate on the retrieval task, we follow the same procedure as in Johnson~\etal\cite{johnson2016densecap} with {our relational captioning data.}
We randomly choose 1000 images from the test set, and from these chosen images, we collect 100 query sentences by sampling four random captions from 25 randomly chosen images.
The task is to retrieve the correct image for each query by matching it with the generated captions.

We compute the ratio of the number of queries, of which the retrieved image ranked within top $k \in \{1,5,10\}$, and the total number of queries (denoted as \emph{R@K}).
We also report the median rank of the correctly retrieved images across all 1000 test images ({The random chance performance is 0.001, 0.005, and 0.01 for R@1, R@5, and R@10 respectively}).
The retrieval results compared with several baselines are shown in Table~\ref{table:retrieval}.
For baseline models \emph{Full Image RNN}, \emph{Region RNN}, and \emph{DenseCap}, we display the performance measured from Johnson~\etal\cite{johnson2016densecap}.
To be compatible, we followed the same procedure of running through random test sets 3 times to report the average results.
Our matching score is computed as follows.
For every test image, we generate 100 region proposals from the RPN followed by NMS.
In order to produce a matching score between a query and a region pair in the image, we compute the probability that the query text may occur from the region pair.
Among all the scores for the region pairs from the image, we take the maximum matching score value as a representative score of the image.
This score is used as the matching score between the query text and the image, and thus the images are sorted by rank based on these computed matching scores.
As shown in Table~\ref{table:retrieval}, the proposed relational captioner outperforms all baseline frameworks. 
{This is meaningful because region pair based method is more challenging than a single region based approaches.}

\begin{table}[t]
	\centering 
    \resizebox{0.8\linewidth}{!}{%
    \begin{tabular}{l|| c|c|c|c}\hline
    &		R@1		&	R@5	&	R@10		& 	Med	\\\hline\hline
    Full Image RNN\cite{karpathy2015deep}	&		0.10	&	0.30	&	0.43	&	13	\\
    Region RNN~\cite{girshick2015fast}&		0.18	&	0.43	&	0.59	&	7	\\
    DenseCap~\cite{johnson2016densecap}		&		0.27	&	0.53	&	{0.67}	&	5	\\\hline
    RelCap (\textbf{\texttt{MTTSNet}})		&\textbf{0.29} &\textbf{0.60}	&\textbf{0.73}&\textbf{4}\\\hline
    \end{tabular}
    }
	\caption{Sentence based image retrieval performance compared to previous frameworks. We evaluate ranking using recall at $k$ (R@$K$, higher is better) and the median rank of the target image (Med, lower is better). 
	\vspace{-4mm}}
    \label{table:retrieval}	
\end{table}


\vspace{-2mm}
\section{Conclusion}
\vspace{-2mm}
We introduce relational captioning, a new notion which requires a model to localize regions of an image and describe each of the relational region pairs with a caption. 
To this end, we propose the MTTSNet, which facilitates POS aware relational captioning. 
In several sub-tasks, we empirically demonstrate the effectiveness of our framework over scene graph generation and the traditional captioning frameworks.
As a way to represent imagery, the relational captioning can provide diverse, abundant, high-level and interpretable representations in caption form.
In this regard, our work may open interesting applications, \eg, natural language based video summarization~\cite{choi2018contextually} may be benefited by our rich representation.
\newline


\noindent{\textbf{Acknowledgements.}}
This work was supported by Institute for Information \& communications Technology Planning \& Evaluation(IITP) grant funded by the Korea government(MSIT) (No.2017-0-01780, The technology development for event recognition/relational reasoning and learning knowledge based system for video understanding)

{\small
\bibliographystyle{ieee}
\bibliography{egbib}
}
\clearpage

\end{document}